\documentclass[runningheads]{llncs}
\usepackage{graphicx}
\usepackage{amsmath,amssymb} 
\usepackage{amsopn}

\DeclareMathOperator*{\argmax}{argmax}

\usepackage{color}
\usepackage[width=122mm,left=12mm,paperwidth=146mm,height=193mm,top=12mm,paperheight=217mm]{geometry}
\usepackage{subfigure}
\usepackage{wrapfig}
\usepackage{comment}
\usepackage{array,multirow,multicol}
\usepackage{algorithm2e}

\def\pathPrefix{}
\usepackage{diagbox, eqparbox, hhline}
\usepackage{url}
\begin{document}
\pagestyle{headings}
\mainmatter

\title{Confidence from Invariance to Image Transformations} 
\titlerunning{Confidence from Invariance to Image Transformations}

\authorrunning{Y. Bahat, G. Shakhnarovich}

\author{Yuval Bahat\textsuperscript{$\dagger$}\thanks{Part of this work was performed while the author was at TTIC} \and Gregory Shakhnarovich \inst{\mathsection}}

\institute{
	\begin{tabular}{cc}
		\textsuperscript{$\dagger$}Dept. of Computer Science and Applied Math,&	\textsuperscript{$\mathsection$}Toyota Technological\\
		Weizmann Institute of Science,&Institute at Chicago,\\
		Rehovot, Israel&Chicago, IL, USA\\
		\email{yuval.bahat@weizmann.ac.il}&\email{gregory@ttic.edu}
	\end{tabular}
}

\maketitle

\begin{abstract}
We develop a technique for automatically detecting the classification errors
of a pre-trained visual classifier. Our method is agnostic to the form
of the classifier, requiring
access only to classifier responses to a set of inputs. We train
a parametric binary classifier (error/correct) on a representation
derived from a set of classifier responses generated from multiple
copies of the same input, each subject to a
different natural image transformation. Thus, we establish a measure of
confidence in classifier's decision by analyzing the invariance of its
decision under various transformations. In experiments with multiple data
sets (STL-10,CIFAR-100,ImageNet) and classifiers, we demonstrate new
state of the art for the error detection task. In addition, we apply
our technique to novelty detection scenarios, where we also
demonstrate state of the art results.

\end{abstract}

\begin{figure}[b]
	\centering
	\includegraphics[width=1\columnwidth]{\pathPrefix 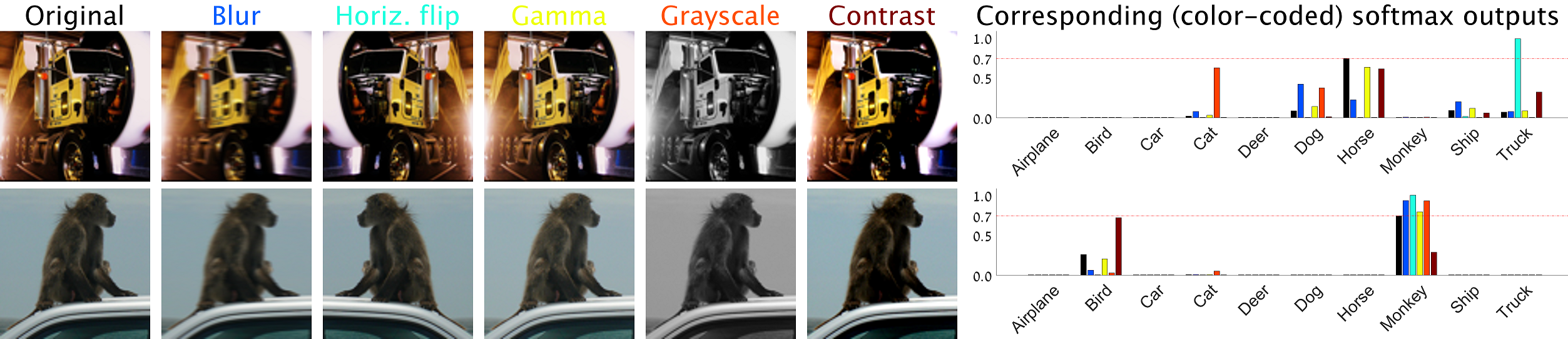}
	
	\caption{\label{fig:transformed_images}\textbf{Correctly and incorrectly classified STL-10 images. }From left to right: The original images, their transformed versions and their corresponding (color-coded) softmax outputs (obtained using the pre-trained classifier of \cite{DistBasedScore_Mendelbaum2017}). While both (original) images obtain the same prediction confidence based on the maximal softmax response (red dashed line), the softmax values of the misclassified (upper) image fluctuate across transformed versions, while those of the correctly classified (lower) image are relatively stable.}
	{\vspace{-0.2cm}}
\end{figure}

\section{Introduction}

Despite rapid and continuing dramatic improvements in the accuracy of
predictors applied to computer
vision, these predictors (image classifiers, object detectors, etc.)
continue to have a non-trivial error rate. For instance, typical error rate
(top-5)
of modern classifiers on ImageNet~\cite{ImageNet} is
on the order of 5\%~\cite{ResNet,huang2017densely}. Error rates increase further when an adversarial process
generates input images designed to ``fool'' the
classifiers~\cite{Ostrich_Szegedy2013,nguyen2015deep}. This observation has led to widespread concerns
regarding robustness of neural network classifiers, and sparked
interest in building various defenses. In this paper, we 
pursue a more general goal:
error detection. More specifically, enabling endowing a
classifier with a \emph{reject option} -- the
possibility of signaling lack of confidence in its ability to correctly classify it.

Our work follows two key ideas. The first is to leverage the rich
signal in the output of a trained classifier to reason about the
likelihood of its prediction being wrong. This approach,
intuitively related to the ``dark knowledge'' ideas~\cite{hinton2015distilling}, goes beyond prior work
which narrowly focused on the score/probability of the predicted
class. The second idea is to leverage signal from the stability of classifier
output under a set of natural image transformations. Intuitively, one
can expect that the corresponding changes in output when the prediction is correct
may differ in a systematic way from such changes when the prediction is
incorrect. For instance, we may expect more invariance to
transformations of a correctly classified input, as demonstrated in Fig.~\ref{fig:transformed_images}.

We convert these two intuitions into a specific model for
error detection: a predictor mapping the collected outputs under a set of
transformations to probability of the original input being
misclassified. This approach achieves results significantly exceeding prior
state of the art for error detection on a number of data sets.


We can also apply this intuition, and our approach, to the closely
related problem of novelty detection. In standard classification
settings, all inputs are assumed to correspond to an available output
class. However, a visual recognition system deployed in the wild is likely to encounter
inputs that belong to \emph{novel} classes, not included in the
training set. In this situation an
error is unavoidable, unless the classifier is endowed with a ``reject
option''.  

An additional scenario where the classifier
may be expected to reject an input is when that input comes from a domain
substantially different from that of the training data (e.g. feeding an image of a digit to a classifier trained to recognize animals). We show that
our approach is applicable, and achieves state of the art results, in these novelty detection scenarios as
well. To the best of our knowledge, we are the first to establish state of the
art in error detection on the ImageNet data set, which is
more realistic and larger than the smaller data sets typically used in
error detection literature.


\section{Related work}
The idea of developing reject option for classifiers has been
discussed in machine learning literature
extensively~\cite{bartlett2008classification,cortes2016boosting,geifman2017selective}. We
borrow some evaluation methodology from this prior work, in
particular~\cite{geifman2017selective}, but our approach is different
from anything proposed there, in particular in our use of image
transformations. 

The body of recent related work can be broadly divided into error detection
methods and novely detection methods. 
A common and well performing baseline method for both tasks is the \emph{Maximal softmax
  Response} (MSR)~\cite{Baseline_Hendrycks2016}, that interprets the maximal output (the one corresponding
to the predicted class) of a softmax classifier as a confidence
score. Thresholding this score is then used to detect
misclassifications or novel class instances. The intuition is that the
output of the original classifier may be poorly calibrated (i.e., will poorly reflect the actual posterior probability $p(\widehat{c}(x)=c|x)$), due to the vagaries
of optimizing a particular surrogate loss such as cross entropy, and
this thresholding step adjusts the
calibration. In~\cite{DistBasedScore_Mendelbaum2017}, calibration is
based on estimating density in feature space corresponding to the last
layer of a deep neural network before softmax. In order to work well, the
approach in~\cite{DistBasedScore_Mendelbaum2017} requires training the
classifier network with a specially modified loss.

In contrast to MSR, we
look beyond the maximum of the posterior distribution, and consider a
much richer family of detectors (a multilayer perceptron, rather than
the simple thresholding in MSR). In contrast to
both MSR and~\cite{DistBasedScore_Mendelbaum2017}, we go beyond
reasoning about scores on the input at hand, and consider the behavior
of the classifier on a set of transformed (perturbed) versions of the input.

A number of methods have been proposed that incorporate reasoning
about stability under perturbations into efforts to improve visual
classification. In~\cite{Dropout_Gal2015} the perturbations consist of applying
stochastic dropout to classifier activations. 
The work in \cite{CounterAdversarialWithManips_Guo2018} proposes to
defend against adversarial attacks
\cite{Ostrich_Szegedy2013,Adversarial_Goodfellow2014} using image
transformations. However, unlike our work, they employ non-natural
transformations, and use those as a pre-processing step rather than a
test-time device. This
requires retraining the classifier to correct for resulting
artifacts. Neither of
these efforts is aimed at error detection, which is our goal.

One key distinction between methods for error detection is their required level
of access to parameters of the underlying classifier. ``White box''
methods~\cite{DistBasedScore_Mendelbaum2017,wang2017safer} require full access, and in fact need
to re-train the classifier to fit their error detection
framework. Other methods are perhaps better characterized as
``gray box'' requiring some degree of access but not full retraining;
this includes the dropout method of~\cite{Dropout_Gal2015}. In contrast, MSR,
and our proposed approach, do not require any access to the classifier
beyond treating it as a ``black box'' which takes in an image and
produces softmax scores for the classes. Thus we can apply our method
to endow a fixed, pre-trained classifier with a robust reject option,
as we demonstrate in Sec.~\ref{sec:exp}.


\section{Confidence from invariance}
The techniques we develop below for error and novelty detection are
applicable to any classifier trained to output softmax scores for the
classes given an image. These techniques require only a ``black box''
level access to the classifier: the ability to feed it an image and
observe the output scores.

\subsection{Error detection from class posteriors}\label{sec:det}
A $k$-way classifier $h$ outputs for an input $x$ a $k$-dimensional
vector of class scores (logits) $\mathbf{s}^h(x)\in\mathbb{R}^k$. Using the
softmax transformation, the scores can be
converted to estimated posterior
distribution over classes $\mathbf{p}^h(x)=h(x)$, where $p_c(x)$ is the
estimated conditional probability of $c\in[k]$ being the class of $x$. The
prediction is made by selecting $\widehat{c}(x)=\argmax_cp_c^h(x)$.
One could use the value $p_{\widehat{c}}^h(x)$ as a measure of
confidence in $h$'s decision on $x$. That's the approach in MSR~\cite{Baseline_Hendrycks2016}. However, it has been shown that
the entire posterior distribution contains information valuable to the
decision making process. This has been called ``dark
knowledge'' in~\cite{hinton2015distilling}. 

We propose to directly exploit this information by training a
binary classifier, mapping either the scores $\mathbf{s}$ or the
posterior $\mathbf{p}$ to the
probability of the input (on which this posterior was calculated)
being misclassified. To this end, we collect a
set of class-labeled examples $(x_i,y_i)$, with the corresponding
scores $\mathbf{s}^h(x_i)$. Using the trained classifier $h$, we now label each score with the
binary \emph{error label} $e_i$, where $e_i=0$ if $y_i=\widehat{c}(x)$
(correct prediction)
and $e_i=1$ otherwise (error). Having constructed such a training set, we can learn
a binary classifier $d$, producing
$d(x)=p\left(e_i=1|\mathbf{s}^h(x_i)\right)$, for instance by
minimizing the cross entropy loss on the $e_i$ labels. To avoid confusion, we
will refer to the original (multi-class) classifier $h$ as
``classifier'', and to the (binary) classifier $d$ predicting whether
$h$ makes an error, as ``detector''. 


One can expect that $h$ will be more
``confident'', and make fewer mistakes on the training data than on
future test data; this is the standard overfitting problem. Therefore,
we will train $d$ on examples not included in training set for
$h$.

\subsubsection{Score representation}The score
vectors $\mathbf{s}^h(x)$ will vary drastically depending on the
estimated class distribution. Our detector needs to learn a mapping from these vectors to the probability of error $d(x)$. To facilitate the learning of this mapping, we make the detector invariant to this variation; We convert $\mathbf{s}^h$ to a canonical representation by
sorting its values. Let $c_1$ be the class with the highest $s_c(x)$,
$c_2$ the second highest, etc. In the sorted score representation
\begin{equation}
\label{eq:sorted_scores}
\tilde{\mathbf{s}}^h(x_i)\,=\,\left[s_{c_1}^h(x),\ldots,s_{c_k}^h(x)\right]
\end{equation}
the class identity
information is lost\footnote{Eliminating the class identity information may seem like a counterintuitive step, but it proves itself empirically in most cases.}, but the shape of the distribution
is retained.

We can now use $\tilde{\mathbf{s}}$
as input to detector $d$, trained to predict $e_i$. In all our
experiments the detector is a multi-layer perceptron (MLP).\footnote{
Note also that we can choose to use the posteriors $\mathbf{p}^h(x)$
rather than the logits $\mathbf{s}$, however it contains strictly less
information than $\mathbf{s}$ due to normalization, and indeed we
found using $\mathbf{p}$ consistently inferior to using $\mathbf{s}$.}

When the number of classes $k$ is large, learning a detector on top of
the full score vector may require too many parameters to learn (or
equivalently too many error/non-error examples). To address this we
can truncate the representation to
$[\tilde{s}_{c_1}^h(x),\ldots,\tilde{s}_{c_{k'}}^h(x)]$ for $k'<k$. The entire process sketch appears inside the green dashed line in Fig.~\ref{fig:system-overview}.

\subsection{Invariance under transformations}
\label{sec:invariance-concept}

The second component of our approach comes from the expectation that a
correct classification decision may/should be invariant to some image
transformations applied to the input, while an incorrect one may be
less stable. More specifically, we argue that the confidence
(here understood as probability of being correct) of prediction $h(x)$
is related to its invariance to a set of ``natural'' image
transformations of $x$. This set includes transformations that
can be expected to occur naturally in realistic imaging
conditions, while not affecting the semantic content of the image. For instance, horizontal flip since the world is largely
laterally symmetric; contrast variation due to environment/sensor
variability; etc.

Given a family of image transformations $\cal T$, We can assess the
degree of invariance for a given classifier and image pair $(h,x)$, by
observing the difference between $\mathbf{p}^h(x)$ and
$\mathbf{p}^h(t(x))$ for $t\in\cal T$, the classifier's output when
fed the original image $x$ vs. $x$ transformed by $t$.


\begin{figure}[t]
	\centering
	\includegraphics[width=1\columnwidth]{\pathPrefix 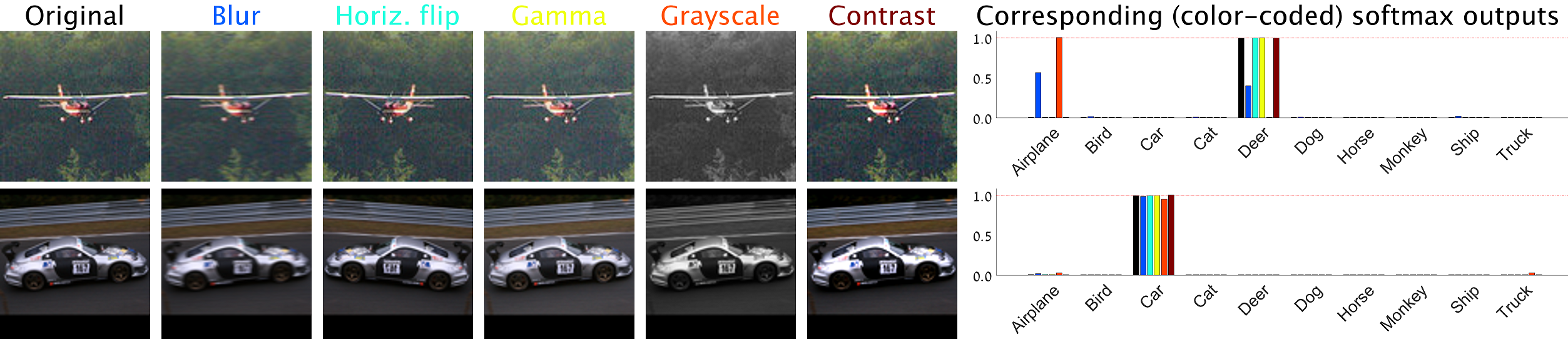}
	
	\caption{\label{fig:transformed_images2}\textbf{Additional examples of correctly and incorrectly classified STL-10 images.}  See caption of Fig.~\ref{fig:transformed_images} for explanation. Note that unlike in Fig.~\ref{fig:transformed_images}, here both images get a very high MSR confidence score (close to 1). Nevertheless, while softmax values of the misclassified (upper) image are inconsistent across transformed versions, those of the correctly classified (lower) image are in concurrence.}
	{\vspace{-0.2cm}}
\end{figure}
As an illustration, Fig.~\ref{fig:transformed_images} and~\ref{fig:transformed_images2} present
two images each, along with their transformed versions under a
set of transformations: horizontal blur, horizontal flip,  gamma
correction, conversion to grayscale, and change in contrast. The right
hand side shows the corresponding softmax outputs for each image
version (including the original). Note that for the correctly
classified images (bottom images, in both figures) the output is fairly consistent across image
versions, while for the misclassified one (top images) the transformations
induce significant changes in the classifier output.
We will now consider computational recipes for converting this
intuition into an error detector.

\subsubsection{Divergence based error detection}
If we expect the correct prediction to come with more stability under
transformation $t$, we can quantify stability by computing a measure of
difference (divergence)
$D\left(\mathbf{p}^h(x)||\mathbf{p}^h(t(x))\right)$ between the two probability distributions. A widely used choice for $D$ is the
Kullback-Leibler divergence $D_{KL}$~\cite{cover2012elements}. Other choices
include the Jensen-Shannon divergence (a symmetrized version of
$D_{KL}$), squared distance, or Kolmogorov-Smirnov divergence.

With either of these choices,
$D\left(\mathbf{p}^h(x)||\mathbf{p}^h(t(x))\right)=0$ when the
classifier's output for image $x$ is completely invariant to the image
transformation $t$, and it increases as $\mathbf{p}^h(t(x))$ diverges more from $\mathbf{p}^{h}(x)$, that is, as the
classifier's prediction on $x$ becomes less invariant under ${t}$.


A straight-forward approach would be to base a detector on
thresholding the divergence under transformation $t$ with threshold $\tau$,
\begin{equation}
  \label{eq:div-det}
  d^t(x)=1\quad\text{if}\quad
  D\left(\mathbf{p}^h(x)||\mathbf{p}^h(t(x))\right)\,\ge\,\tau.
\end{equation}

The approach as developed thus far considers one transformation at a
time. We conjecture (and empirically confirm this conjecture in Sec.~\ref{sec:exp})  that additional power could be
derived from jointly considering multiple transformations. Rather than
pursue heuristic combination rules for a divergence-based detector, we
directly extend the parametric detector framework in
Sec.~\ref{sec:det} to invariance-based scenario with multiple transformations.

\begin{figure}[t]
	\centering
	\includegraphics[width=1\columnwidth]{\pathPrefix 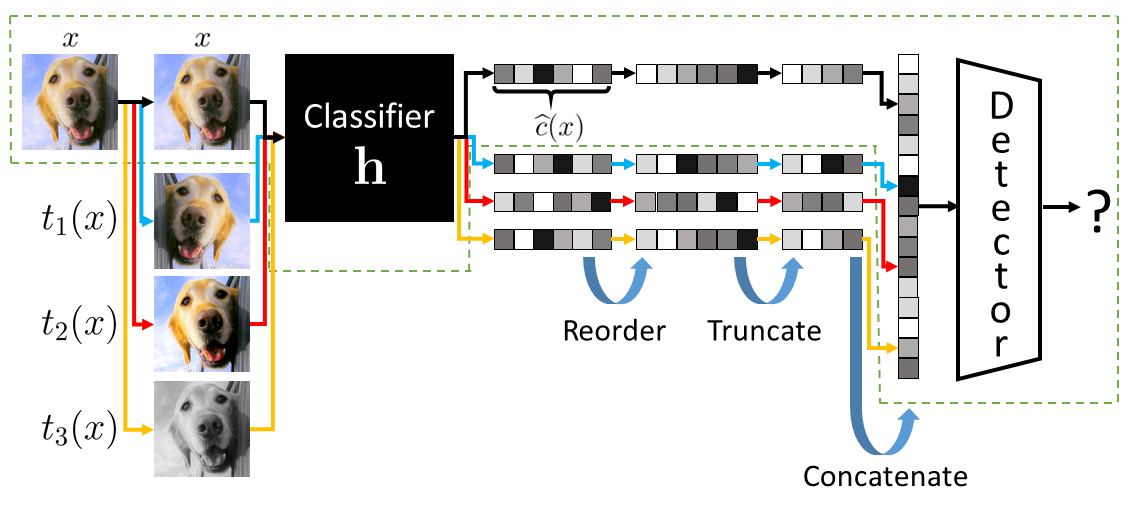}
	
	\caption{\label{fig:system-overview}\textbf{Method overview.}
          Given an image $x$ and a pre-trained classifier $h$, we feed
          $x$ and several (three, in this example) natural
          transformations of it into $h$. We jointly re-order all
          resulting score vectors so that the score vector for $x$
          (the original image) is in descending order. We then
          truncate the score vectors to retain only the $k'$ first
          scores and concatenate them to yield the input to our
          detector. The green dashed line delineates our approach when used without the transformations (denoted by MLP in Sec.~\ref{sec:exp}).}
	{\vspace{-0.2cm}}
\end{figure}

\subsection{Combining confidence and invariance}
\label{sec:our_detector}
Given a set of transformations $\mathcal{T}=\{t_1,\ldots,t_m\}$ and the fixed classifier $h$,
we can compute, for an image $x$, the set of output scores
$\mathbf{s}^h(t_0(x)),\mathbf{s}^h(t_1(x)),\ldots,\mathbf{s}^h(t_m(x))$,
where the ``null transformation'' $t_0$ is identity (i.e., the
original input).

We then jointly order all score vectors $\mathbf{s}^h(t_j(x)),~t_j\in\cal T$ by sorting the scores of the original input;
finally, all the sorted scores are truncated
by keeping only the first $k'\le k$ values. Conceptually, this forms a
re-ordered, truncated $m+1\times k'$ matrix of scores.
Please see Fig.~\ref{fig:system-overview}, depicting this process for $m=3$, $k=6$ and $k'=4$.

\subsection{Novelty detection}\label{sec:novelty}
When the true label of an image fed to a classifier lies outside the
classifier's $k$ possible classes, the predicted class is guaranteed
to be incorrect. A mechanism to detect such incidents is called
novelty detector. In this work, we refine the notion of novelty into
two scenarios, and use our method to detect each of them. A crucial
distinction from the error detection task above is that here, we can
not possess labeled training examples covering the class distribution of novel images.

\subsubsection{Novel Domains}
\begin{figure}[b]
	\centering
	\includegraphics[width=1\columnwidth]{\pathPrefix 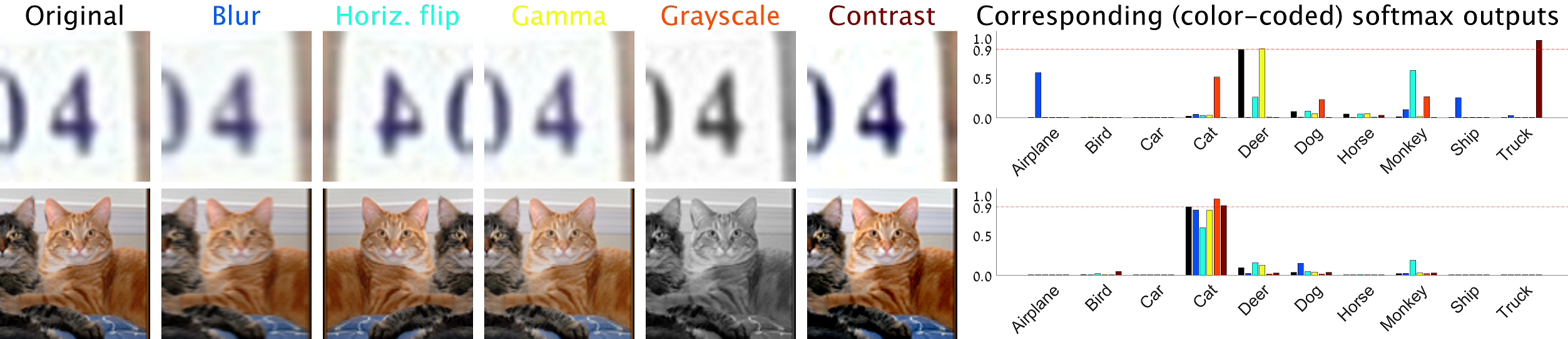}
	
	\caption{\label{fig:transformed_images_novelty}\textbf{Novel (misclassified) image from SVHN and correctly classified STL-10 image.}  See caption of Fig.~\ref{fig:transformed_images} for explanation.}
	{\vspace{-0.2cm}}
\end{figure}

The first novelty scenario (previously explored in, e.g.,
\cite{DistBasedScore_Mendelbaum2017,Baseline_Hendrycks2016}) is the
domain novelty (also known as \emph{out-of-distribution (OOD)}
novelty), where a classifier trained on a certain domain (e.g. STL-10
dataset of objects) is fed with images from a different domain
(e.g. the SVHN \cite{SVHN_Netzer2011} dataset of street numbers, see top row of Fig.~\ref{fig:transformed_images_novelty}). 
To detect OOD images, we use the same framework explained in
Sec.~\ref{sec:our_detector}. We assume that we have access to a data
set of images from a domain different from what the classifier was
trained on, which can generally be different from the domains that may be
encountered in the future. For instance, given a hand-written digit
classifier, we can train the error detector by using images of objects to simulate novel inputs. We can then use this detector to detect images from other novel domains (e.g. animals).

\subsubsection{Novel Classes}
The second novelty scenario we discuss here is more realistic. After a
classifier is trained to distinguish between $k$ classes and is
deployed, it will eventually encounter input images that come from
classes beyond those $k$ (but still from the same domain). In
this scenario, novelty detector must identify images from the
unfamiliar classes, both to prevent errors and, in some applications,
to provide a mechanism for active learning (soliciting labels for
examples of unfamiliar classes). Our method can be employed for this case too.

To overcome the training examples problem in this case, we take a
slightly different approach, and use a hold-out partition of the
original training set for the classifier $h$. Specifically, we hold
out a subset of classes, say $\tilde{k}< k$, and train a classifier $\tilde{h}$
(with the same architecture) on the remaining classes $c\le \tilde{k}$. We then
train an error/novelty detector as in the OOD scenario, but using
examples from the
heldout classes as OOD examples. Once the detector $d$ is trained, we
can apply it to detect novelty for the original, full-set classifier
$h$. We elaborate on this procedure in Sec.~\ref{sec:exp_novel_classes}.

\section{Experiments}
\label{sec:exp}

In our experiments we use three primary data sets: STL-10~\cite{STL10}, CIFAR-100~\cite{CIFAR100}, and
ImageNet~\cite{ImageNet}, as base data sets on which the visual
classifier (whose errors we will want to detect) is trained. Each of these is roughly an order of magnitude larger than the previous
one in terms of both number of examples and the number of
classes. This is to confirm that the proposed method generalized
across such parameters of the data set/classification task.

We
demonstrate results with different classifiers, including a
competitive Inception-ResNet classifier. Our experiments include both
the basic error detection task and multiple novelty detection scenarios. We compare, to the extent
possible, to previously proposed methods, across a range of meaningful
metrics; in contrast to image classification, there is no clear single
way to evaluate error detection or novelty detection.

\subsection{Evaluation Measures}
We use the following measures to evaluate and compare performance.

\noindent\textbf{Area Under Receiver Operating Characteristic Curve}
(AUROC) relies on the ROC curve (true-positive vs. false positive
rates for all possible thresholds on a detector's output). The area
under the ROC curve is invariant to the ``polarity'' of our detection
task - whether we detect correct or incorrect classifications -
because the curves for these two definitions are symmetric. However,
the AUROC is dependent on the ratio of incorrect vs. correct
classifications, hence this ratio in the test set should be maintained constant when comparing different detectors.

\noindent\textbf{Coverage vs. Accuracy Curve} (CAC) that relates to using the reject option for classification. Each point $(coverage,accuracy)$ in the curve reflects the anticipated $accuracy$ when using detector $d$ to keep only the $coverage$ portion of classifications corresponding to the lowest error detection score $d(x)$. Similar to AUROC, we also use the area under the CAC (AUCAC) to evaluate performance. 


\subsection{Error detection: Experimental setup}
Throughout our experiments we use a fully connected MLP detector that
has two hidden layers of width 70, each followed by  RELU
nonlinearity, and a batch normalization layer. We train our detector
using asymmetric cross-entropy loss, reweighting examples to correct
for the binary class (correct / incorrect) imbalance, and with dropout
probability $0.5$ in both hidden layers.

We used the following set of natural image transformations $\mathcal{T}$ throughout our experiments:

\noindent\emph{Horizontal flip:} Flipping the image on the horizontal axis.\\
\noindent\emph{Horizontal blur:} Blurring the image with a horizontal blur kernel. We used a magnitude of 3 pixels in our implementation.\\
\noindent\emph{Converting to gray-scale:} Overriding all three channels ($R,G,B$) of each image pixel with $0.299\times R+0.587\times G+0.114\times B$.\\
\noindent\emph{Contrast enhancement:} Increasing image contrast by $1.3$.\\
\noindent\emph{Gamma correction:} Raising each image pixel to the power of $0.85$.

%

To avoid the problem of fitting our detector to training-set images
outputs, we train it using validation-set images. To this end, we
split the originally given validation set to two subsets, and use them
as training and validation sets for our detector.\footnote{The subset
  assignment for each data-set is consistent across all our
  experiments, and will be made publicly available,
  along with our code.} As explained in Sec.~\ref{sec:our_detector},
when using an MLP, we reorder and clip to length $k'$ the output
vectors $\mathbf{s}^h(t_j(x))$ corresponding to each image $t_j(x)$,
before concatenating them into a single input vector for our
detector. We use $k'=5,10,20$ when applying our method to the STL-10,
CIFAR-100 and ImageNet data-set, respectively.

In training error and novelty detectors, we use data augmentation,
by randomly applying horizontal flip and random brightness
and contrast adjustments (and for ImageNet, also a random crop).
When constructing the MLP+Invariance representations, the (fixed) transformations in $\mathcal{T}$ are applied to an input image \emph{after} it has been obtained using the data
augmentation pipeline.

\begin{table}[b]
	\begin{center}
		\begin{tabular}{|c||c||c|c|c|c|c||c|}
			\hline
			\diagbox[innerwidth=3.5cm,trim=r]{Detector}
			{Transformation}& Original&Flip & Gamma & Contrast &Blur & Gray scale & ALL\\

			\hline
			MSR \cite{Baseline_Hendrycks2016}&0.807&-&-&-&-&-&-\\
			Ours, KL divergence&-&0.809&0.787&0.798&0.797&0.708&-\\
			Ours, MLP&0.813&0.834&0.821&0.824&0.825&0.822&0.846\\
			\hline
		\end{tabular}
	\vspace{5pt}
	\caption{\label{tab:error_detect_transformations}\textbf{Error
            detection using different image transformations, STL-10.}
          AUROC values of detectors based on scores of different image
          transformations. Middle row values correspond to directly
          using the KL-divergence score, while bottom row values are
          obtained using MLP on top of the logits. All detectors are applied on the \emph{Regular} classifier of \cite{DistBasedScore_Mendelbaum2017}. }
	\end{center}
\end{table}

\subsection{Error detection: results}

\noindent\textbf{STL-10}
We evaluated the performance gains of the different components of our method using a pre-trained STL-10 image classifier (trained by \cite{DistBasedScore_Mendelbaum2017}) with $70\%$ accuracy, and present the result in table~\ref{tab:error_detect_transformations}. 
We followed Eq.~\ref{eq:div-det} and directly used the
KL-divergences between
classifier outputs corresponding to the original images vs. one of the
transformations in $\mathcal{T}$, as an error detection
mechanism. This by itself (middle row) gave comparable performance to
MSR.  
Using other divergence measures mentioned in Sec.~\ref{sec:invariance-concept} gave results which were consistent with those of KL-divergence.

We omit those divergences from further discussion.

\begin{wrapfigure}[14]{r}{0.3\textwidth}
	\centering
	\includegraphics[width=0.3\textwidth]{\pathPrefix 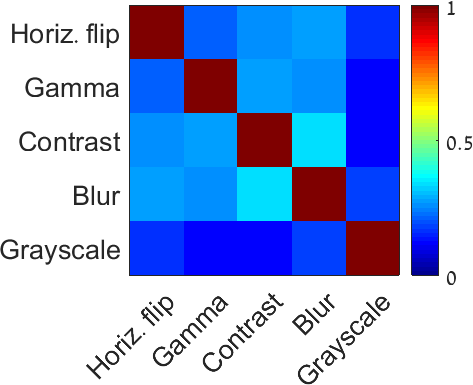}
	
	\caption{\label{fig:T-correlations}Correlation between KL-divergences for different $t\in\mathcal{T}$.}
\end{wrapfigure}

  Next, we evaluate the power of combining several image
  transformations. As a diagnostic experiment, we compute correlation coefficients
  between KL-divergence values (between posteriors for the original
  input and the posterior for the result of transformation $t$)
  computed for different $t\in\mathcal{T}$. As shown in
  Fig.~\ref{fig:T-correlations}), many of these correlations are low,
  indicating that the information provided by outputs on different
  $t$s is not completely redundant.

The bottom row of table~\ref{tab:error_detect_transformations}
presents the performance achieved when using an MLP and feeding it
with the classifier's output corresponding to (1) the original image
alone (left column), (2) the original image and one transformed image
(middle columns) and (3) all image versions (right column). This
exploits the full power of our method and achieves a significant
increase in error detection performance (approximately 4 points gain in AUROC compared
to MSR or KL-divergence).

\begin{figure}[!t]
	\centering
	\begin{minipage}{0.48\textwidth}
		\centering
		\includegraphics[width=1\textwidth]{\pathPrefix 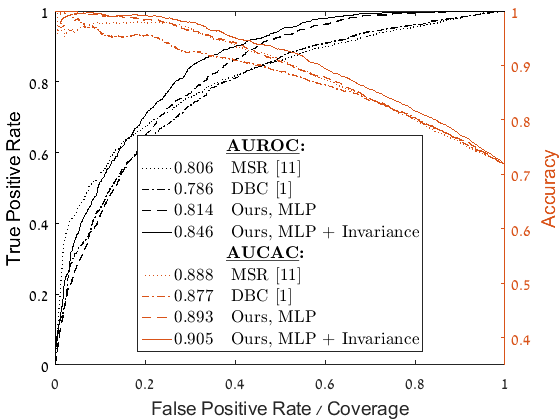}
	\end{minipage}\hfill
	\begin{minipage}{0.48\textwidth}
		\centering
		\includegraphics[width=1\textwidth]{\pathPrefix 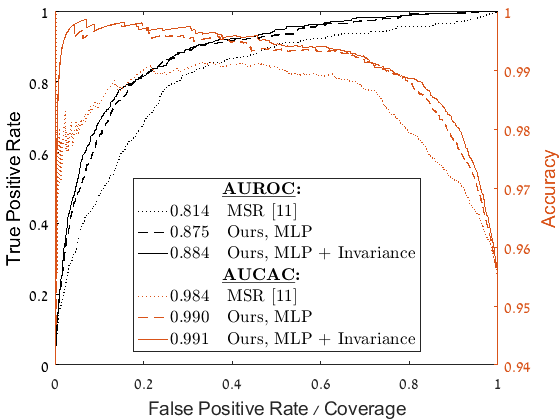}
	\end{minipage}
		\caption{\label{fig:ROC_CAC_curves}\textbf{Error detection on the STL-10 (left) and ImageNet top-5 (right) classification tasks.} ROC and CAC corresponding to error detection using MSR \cite{Baseline_Hendrycks2016} and our method, with and without using invariance to image transformations. For STL-10, we also plot the curves for DBC \cite{DistBasedScore_Mendelbaum2017}, and leave out the plot for MC-dropout \cite{Dropout_Gal2015}, which performs similarly to MSR (see table~\ref{tab:error_datect_methods_Mendelbaum_AUROC_clsfr}). Values in the legend correspond to Area Under Curves.}
\end{figure}

More extensive comparison of our method to MSR
\cite{Baseline_Hendrycks2016}, MC-dropout \cite{Dropout_Gal2015} and
the SOTA method by Mandelbaum and Weinshall
\cite{DistBasedScore_Mendelbaum2017} on STL-10 are shown in
Tables~\ref{tab:error_datect_methods_Mendelbaum_AUROC_clsfr}
and~\ref{tab:error_datect_methods_Mendelbaum_AUCAC_clsfr}. Our
evaluation included pre-trained classifiers used
in~\cite{DistBasedScore_Mendelbaum2017}\footnote{Following
  \cite{DistBasedScore_Mendelbaum2017}, we pre-processed the original
  and transformed images by performing global contrast minimization
  and ZCA whitening.}, that differ in the loss function they utilized;
The \emph{Regular} classifier used only the cross entropy
loss. \emph{Dist.} and \emph{AT} both employ the cross-entropy loss,
but augment it with a features embedding term or the adversarial loss
term of \cite{Adversarial_Goodfellow2014}, respectively. The latter
two loss functions require retraining the classifier to improve the
performance of DBC, and thus are by design favorable to DBC (and in
that sense do
not satisfy our desired black box treatment of the classifier).


The resulting ROC and CAC (corresponding to the `Regular' classifier)
appear in Fig.~\ref{fig:ROC_CAC_curves} (left). AUROC and AUCAC values
for all three classifiers are presented in
tables~\ref{tab:error_datect_methods_Mendelbaum_AUROC_clsfr}
and~\ref{tab:error_datect_methods_Mendelbaum_AUCAC_clsfr},
respectively. \begin{table}[b]
	\begin{center}

		\begin{tabular}{|c||c|c|c|c|c|c|}
			\hline
			\multirow{2}{*}{}& \multicolumn{3}{c}{STL-10} & \multicolumn{3}{|c|}{CIFAR-100}\\\cline{2-7}
			& Regular & Dist. & AT & Regular & Dist. & AT\\
			\hline
			Accuracy (\%)&71.8&71.9&70&58.8&57.8&59.1\\
			\hline
			MSR \cite{Baseline_Hendrycks2016} &0.806&0.772&0.813&0.834&0.83&0.842\\
			MC-dropout\cite{Dropout_Gal2015} &0.803&0.78&0.809&0.834&0.83&0.847\\			DBC \cite{DistBasedScore_Mendelbaum2017} &0.786&0.816&0.866&0.782&0.852&0.858\\
			Ours, MLP &0.814&0.789&0.817&0.848&0.833&0.846\\
			Ours, MLP+Inv. &\textbf{0.846}&\textbf{0.836}&\textbf{0.868}&\textbf{0.864}&\textbf{0.857}&\textbf{0.869}\\
			\hline
		\end{tabular}
		\vspace{3pt}
	\caption{\label{tab:error_datect_methods_Mendelbaum_AUROC_clsfr}\textbf{Error detection AUROC on pre-trained classifiers of \cite{DistBasedScore_Mendelbaum2017}.} Comparing our method (bottom row) to MSR \cite{Baseline_Hendrycks2016}, MC-Dropout \cite{Dropout_Gal2015} and the method of Mendelbaum and Weinshall \cite{DistBasedScore_Mendelbaum2017}, when tested on classifiers trained by \cite{DistBasedScore_Mendelbaum2017}.}
	\vspace{-1.2cm}
	\end{center}
\end{table}

\begin{table}
	\begin{center}
		\begin{tabular}{|c||c|c|c|c|c|c|}
			\hline
			\multirow{2}{*}{}& \multicolumn{3}{c}{STL-10} & \multicolumn{3}{|c|}{CIFAR-100}\\\cline{2-7}
			& Regular & Dist. & AT & Regular & Dist. & AT\\
			\hline
			Accuracy (\%)&71.8&71.9&70&58.8&57.8&59.1\\
			\hline
			MSR \cite{Baseline_Hendrycks2016} &0.888&0.862&0.883&0.825&0.815&0.832\\
			MC-dropout\cite{Dropout_Gal2015} &0.89&0.869&0.883&0.827&0.815&0.834\\			DBC \cite{DistBasedScore_Mendelbaum2017} &0.877&0.892&\textbf{0.904}&0.772&0.824&0.833\\
			Ours, MLP &0.893&0.874&0.887&0.832&0.817&0.834\\
			Ours, MLP+Inv. &\textbf{0.905}&\textbf{0.9}&0.903&\textbf{0.839}&\textbf{0.827}&\textbf{0.843}\\
			\hline
		\end{tabular}
		\vspace{3pt}		
	\caption{\label{tab:error_datect_methods_Mendelbaum_AUCAC_clsfr}\textbf{Error detection AUCAC on pre-trained classifiers of \cite{DistBasedScore_Mendelbaum2017}.} Different measure for the same experiment as in table~\ref{tab:error_datect_methods_Mendelbaum_AUROC_clsfr}.\vspace{-3em}}
	\end{center}
\end{table}

Our method achieves the best performance on all three classifiers (except for one metric in one experiment). Note that while our detector can be used on any given, pre-trained, classifier, it achieves SOTA performance even on classifiers that were especially modified by \cite{DistBasedScore_Mendelbaum2017} (the AT and Dist. configurations).

\noindent\textbf{CIFAR-100} Next, we performed an evaluation on the CIFAR-100 dataset. We followed the same experiment as before, and used the three pre-trained classifier of the DBC method \cite{DistBasedScore_Mendelbaum2017}. A comparison of AUROC and AUCAC values is presented in tables~\ref{tab:error_datect_methods_Mendelbaum_AUROC_clsfr} and~\ref{tab:error_datect_methods_Mendelbaum_AUCAC_clsfr}, respectively. Here too, our method achieves SOTA performance on all classifiers, including those modified by DBC.

\noindent\textbf{ImageNet} Finally, we evaluated our method's
performance when applied to the ILSVRC-2012 ImageNet classification
task. We used a pre-trained Inception-ResNet-v2 model \cite{ResNet}
that achieves $81\%$ and $95.5\%$ top-1 and top-5 accuracies,
respectively. As before, we used part of ImageNet's validation set to
train our detector, and evaluated its performance on the remaining
part ($20\%$ in this experiment). Fig.~\ref{fig:ROC_CAC_curves}
(right) presents ROC and CAC for detecting errors in the top-5
classification task. We compare curves corresponding to our method
(with and without employing transformations) and MSR\footnote{We tried
  adopting MSR to the top-5 task by summing over the top 5 maximal
  soft-max responses instead of just taking the highest one, but it
  yielded inferior results.}. Table~\ref{tab:error_detect_ImageNet}
presents the corresponding AUROC and AUCAC values. Due to the scale of
ImageNet, we did not compare to ``white box'' methods that require
either extensive re-training (DBC) or extensive computation
(MC-dropout).

\begin{table}[!thb]
	\begin{center}
		\begin{tabular}{|c||c|c|}
			\hline
			& Top-1 & Top-5\\\cline{2-3}
			Accuracy (\%)&81&95.5\\
			\hline
			MSR \cite{Baseline_Hendrycks2016} &0.842/0.936&0.806	/0.983\\
			Ours, MLP &0.866/0.948&0.877/\textbf{0.991}\\
			Ours, MLP+Inv.&\textbf{0.875}/\textbf{0.952}&\textbf{0.884}/\textbf{0.991}\\
			\hline
		\end{tabular}
		\vspace{5pt}
	\caption{\label{tab:error_detect_ImageNet}\textbf{Error detection performance on a pre-trained Inception-ResNet-v2 ImageNet classifier of \cite{ResNet}.} AUROC/AUCAC values of MSR compared to our method, with (bottom row) and without (middle row) employing invariance to image transformations.\vspace{-3em}}
	\end{center}
\end{table}


\subsection{Novelty detection}
We evaluated the performance of our method when used for detecting novelty of both types described in Sec.~\ref{sec:novelty}, namely novel domain (OOD) and novel classes. Prior to performing the evaluation, we had to decide how to treat cases of non-novel images that are misclassified by the classifier (i.e. errors, in the context of error-detection). We took an operational point of view and chose to treat these cases as novel (hence a novelty detector should detect them), rather than ignore them.
\subsubsection{Domain novelty: setup}
We fed the three pre-trained STL-10 classifiers of DBC (\emph{Regular, Dist.} and \emph{AT}) with images from either the CIFAR-100 or the SVHN data-sets. In both cases, half the images came from the original data-set (STL-10) and the other half from the novel data-set (CIFAR-100 or SVHN).

\begin{algorithm}[!h]
	\begin{enumerate}
		\item Define ${\cal C}_N\subset{\cal C}$ of size $S_N=|{\cal C}_N|$, simulating novel, unknown classes. Denote by ${\cal C}_F$ its complementary subset of familiar, known classes ${\cal C}_F={\cal C}\setminus{\cal C}_N$.
		\item Train a classifier $h$ using only images from classes $\in {\cal C}_F$
		\item Set $T_F=\{\}$, a training set containing output scores corresponding to images and their transformations.
		\item \For{All possible subsets $\tilde{{\cal C}}_N\subset{\cal C}_F$ of size $S_N$ (and their complementaires, $\tilde{{\cal C}}_F$)}{
			\begin{description}
				\item[(a)] Train a classifier $\tilde{h}$ using only images from classes $\in \tilde{{\cal C}}_F$
				\item[(b)] Feed $\tilde{h}$ with images from all familair classes (${\cal C}_F$) and their transformed versions (Note that $S_N$ of the classes in ${\cal C}_F$ are unkonwn to $\tilde{h}$). Add the corresponding output logits of $\tilde{h}$ to the training set $T_F$.
			\end{description}
		}
		\item Train an MLP novelty detector using $T_F$.
		\item Evaluate the detector performance on classifier $h$, by feeding it with images from all classes $\in{\cal C}$ (and their transformed versions).
                \end{enumerate}
                \label{alg:pairs}\caption{The training procedure used for
                   class novelty experiments on STL-10. $\mathcal{C}$
                   is the complete set of classes in the data set.}
               \end{algorithm}

\begin{table}[!b]
	\begin{center}
		\begin{tabular}{|c||c|c|c|c|c|c|}
			\hline
			\multirow{2}{*}{}& \multicolumn{3}{c}{SVHN is novel} & \multicolumn{3}{|c|}{CIFAR-100 is novel}\\\cline{2-7}
			& Regular & Dist. & AT & Regular & Dist. & AT\\
			\hline
			Accuracy (\%) &35.9/50&35.9/50&35.0/50&35.9/50&35.9/50&35.0/50\\
			\hline
			MSR \cite{Baseline_Hendrycks2016} &0.818&0.798&0.745&0.802&0.775&0.846\\
			DBC \cite{DistBasedScore_Mendelbaum2017} &0.813&0.854&0.893&0.783&0.849&0.883\\
			MC-dropout \cite{Dropout_Gal2015} &0.819&0.808&0.743&0.804&0.792&0.846\\
			Ours &0.825&0.847&0.902&0.818&0.837&0.879\\
			Ours, Cross-train &\textbf{0.898}&\textbf{0.889}&\textbf{0.914}&\textbf{0.86}&\textbf{0.874}&\textbf{0.907}\\
			\hline
		\end{tabular}
		\vspace{5pt}
	\caption{\label{tab:novelty_domains}\textbf{Novelty detection, across domains.} The STL-10 pre-trained classifiers of \cite{DistBasedScore_Mendelbaum2017} were fed with images from SVHN and CIFAR100 datasets. Values represent AUROC on validation sets constituting 50\% novel images (hence the maximum accuracy is 50\%). Misclassified STL-10 images were considered novel for this experiment. \emph{Ours, Cross-train} is our detector when trained on novel images from the non-participating domain - CIFAR-100 for the `SVHN is novel' experiment, and vice versa.}
	\end{center}
\end{table}

\subsubsection{Domain novelty: results}
Table~\ref{tab:novelty_domains} presents AUROC results
for detecting novel images.
Note that in this case, training our detector on familiar domain images only (denoted by `Ours') yields inferior results. This problem is solved, however, by augmenting our detector training set with images from a domain with which the classifier at hand is unfamiliar (`Ours, Cross-train').

\begin{table}[!h]
	\begin{center}
		\begin{tabular}{|c||c||c|c|c|}
			\hline
			\multirow{2}{*}{Novel classes (${\cal C}_N$)}&Classifier's&\multirow{2}{*}{MSR \cite{Baseline_Hendrycks2016}}&\multirow{2}{*}{Ours, MLP}&\multirow{2}{*}{Ours, MLP+Inv.}\\
			& accuracy (\%)&&&\\
			\hline
			(0)-Airplane, (1)-Bird &57.4&0.732/0.646&0.734/0.648&\textbf{0.774/0.68}\\
			(2)-Car, (3)-Cat &54.5&0.717/0.622&0.724/0.62&\textbf{0.78/0.659}\\
			(4)-Deer, (5)-Dog&57.9&0.718/0.65&0.722/0.652&\textbf{0.772/0.685}\\
			(6)-Horse, (7)-Monkey&54.5&0.724/0.627&0.722/0.63&\textbf{0.769/0.662}\\
			(8)-Ship, (9)-Truck&56.3&0.655/0.553&0.637/0.548&\textbf{0.691/0.577}\\
			(0)-Airplane, (5)-Dog&59.1&0.724/0.655/&0.73/0.663&\textbf{0.768/0.683}\\
			(3)-Cat, (6)-Horse&55&0.725/0.636&0.723/0.635&\textbf{0.776/0.668}\\
			(4)-Deer, (8)-Ship&57&0.706/0.63&0.709/0.639&\textbf{0.762/0.666}\\
			(2)-Car, (7)-Monkey&60.8&0.708/0.648&0.712/0.654&\textbf{0.749/0.67}\\
			(1)-Bird, (9)-Truck&61.6&0.737/0.677&0.746/0.685&\textbf{0.782/0.705}\\
			\hline\hline
			Average&57.4&0.715/0.634&0.716/0.637&\textbf{0.762/0.666}\\
			STD&2.5&0.023/0.033&0.03/0.035&0.027/0.034\\
			\hline
		\end{tabular}
		\vspace{5pt}
	\caption{\label{tab:novelty_classes}\textbf{Novelty detection, within domain (novel classes).} Each row corresponds to different classes chosen to simulate novelty. From left to right: Chosen classes, classifier's accuracy on the 8 non-novel classes and AUROC/AUCAC values for MSR and our method, without and with employing classifier's invariance to image transformations. Average and STD values appear in the bottom two rows.}
	\vspace{-1.2cm}
	\end{center}
\end{table}

\subsubsection{Class novelty: setup}
\label{sec:exp_novel_classes}
To simulate this scenario and evaluate our detector's performance in
this case, we use the STL-10 data-set with its 10 classes ($|{\cal
  C}|=10$) and follow the procedure in Alg.~\ref{alg:pairs}.

In this experiment we used a simple architecture (available on-line in \cite{CIFAR10_tutorial}) as our classifiers $h$ and $\tilde{h}$. To this end, we downscaled the 96x96 pixels STL-10 images to 32x32 pixels before feeding them into our classifiers. This architecture has 2 convolutional blocks, each constituting a max-pool and a batch-normalization layer, followed by two fully connected layers. We trained each of the classifiers $h$ (for a given choice of ${\cal C}_N$) and $\tilde{h}$ (for all classes subsets $\tilde{{\cal C}}_F$ induced by this chosen ${\cal C}_N$) for two hours (in parallel), reaching the accuracy of approximately $55\%-60\%$ on non-novel classes.

\subsubsection{Class novelty: results}
We repeated the experiment in Sec.~\ref{sec:exp_novel_classes} for ten, arbitrarily chosen pairs of classes that simulate novelty. AUROC and AUCAC values are reported in table~\ref{tab:novelty_classes}, along with average and standard deviation values. 

Using our method without the invariance part achieves comparable results to the other ``black-box'' method, MSR. However, exploiting the full power of our method brings a major leap in performance in this realistic novelty scenario.

\section{Conclusion}\label{sec:conclusion}

We have presented a new approach to error and novelty detection in
visual classification, based on analysis of stability of classifier's
output under a set of natural input transformations. We use a multi-layer perceptron detector whose input is the output scores of a visual classifier. In contrast to many previous efforts, our approach only requires a black box level access to the classifier, and thus can be applied to any off
the shelf, pre-trained classifier. We demonstrate new state of the art achieved by our method on both error detection and novelty (novel domain or novel class) detection on a number of data sets, including ImageNet paired with a modern ResNet-based classifier.
We believe the notion of classifier's invariance to natural image transformations can be further exploited in the future, for example by incorporating it into the training procedure of visual classifiers, in an attempt to yield accuracy gains.

\bibliographystyle{splncs}
\bibliography{\pathPrefix ms}
\end{document}